%% file: main.tex
\documentclass[10pt,twocolumn,letterpaper]{article}

\usepackage{cvpr}              

\input{preamble}

\definecolor{cvprblue}{rgb}{0.21,0.49,0.74}
\usepackage[pagebackref,breaklinks,colorlinks,allcolors=cvprblue]{hyperref}

\def\paperID{31257}
\def\confName{CVPR}
\def\confYear{2026}

\title{Benchmarking Layout-Guided Diffusion Models through Unified Semantic-Spatial Evaluation in Closed and Open Settings}

\author{
Luca Parolari\thanks{Corresponding author. Email: luca.parolari@studenti.unipd.it} 
\qquad Nicla Faccioli 
\qquad Lamberto Ballan
\\
Department of Mathematics, University of Padova, Padova, Italy\\
}

\begin{document}

\maketitle

\begin{abstract}
Evaluating layout-guided text-to-image generative models requires assessing both semantic alignment with textual prompts and spatial fidelity to prescribed layouts.
Assessing layout alignment requires collecting fine-grained annotations, which is costly and labor-intensive. 
Consequently, current benchmarks rarely provide comprehensive layout evaluation and often remain limited in scale or coverage, making model comparison, ranking, and interpretation difficult.
In this work, we introduce a closed-set benchmark (\csb{}) designed to isolate key generative capabilities while providing varying levels of complexity in both prompt structure and layout. 
To complement this controlled setting, we propose an open-set benchmark (\osb{}) that evaluates models using real-world prompts and layouts, offering a measure of semantic and spatial alignment in the wild.
We further develop a unified evaluation protocol that combines semantic and spatial accuracy into a single score, ensuring consistent model ranking.
Using our benchmarks, we conduct a large-scale evaluation of six state-of-the-art layout-guided diffusion models, totaling 319,086 generated and evaluated images.
We establish a model ranking based on their overall performance and provide detailed breakdowns for text and layout alignment to enhance interpretability.
Fine-grained analyses across scenarios and prompt complexities highlight the strengths and limitations of current models.
Code is available at \href{https://github.com/lparolari/cobench}{github.com/lparolari/cobench}.
\end{abstract}

\section{Introduction}
\label{sec:intro}

\begin{table}
\centering
\footnotesize
\begin{tabular}{lccccc}
\toprule
Benchmark & Venue & \#Scenario & \#Instr. & Prompt & Layout \\
\midrule
DrawBench~\citep{DBLP:conf/nips/SahariaCSLWDGLA22} & NeurIPS'22 & 4 & 200 & T & \xmark \\
ABC-6K~\citep{DBLP:conf/iclr/FengHFJANBWW23} & ICLR'23 & 1 & 3.2k & E & \xmark \\
CC-500~\citep{DBLP:conf/iclr/FengHFJANBWW23} & ICLR'23 & 1 & 500 & T & \xmark \\
HSR-Bench~\citep{DBLP:conf/iccv/Bakr0SKLE23} & ICCV'23 & 13 & 45k & E & \xmark \\
TIFA v1.0~\citep{DBLP:conf/iccv/HuLKWOKS23} & ICCV'23 & 12 & 4k & T & \xmark \\
T2I-CB~\citep{DBLP:conf/nips/HuangSXLL23} & NeurIPS'23 & 6 & 6k & T & \xmark \\
ConceptMix~\citep{DBLP:conf/nips/WuYHRA24} & NeurIPS'24 & 8 & n/a & L & \xmark \\
Layout-Bench~\citep{DBLP:conf/cvpr/0001LYGWB22} & CVPR'24 & 1 & 8k & T+L & \cmark \\
Hico-7K~\cite{DBLP:conf/nips/ChengMwLMWLY24} & NeurIPS'24 & 2 & 7k & E & \cmark \\
LLM Blueprint~\cite{DBLP:conf/iclr/GaniBN0W24} & ICLR'24 & 1 & n/a & L & \xmark \\
LLM-grounded~\cite{DBLP:journals/tmlr/LianLYD24} & TMLR'24 & 4 & 400 & T+L & \xmark \\
CreatiLayout~\cite{DBLP:journals/corr/abs-2412-03859} & ICCV'25 & 1 & 5k & E & \cmark \\
7Bench~\citep{izzo20257bench} & ICIAP'25 & 7 & 224 & T & \cmark \\
\textbf{Ours} & CVPRF'26 & \textbf{7+1} & \textbf{6.6k} & \textbf{T+L+E} & \textbf{\cmark} \\
\bottomrule
\end{tabular}
\caption{Comparison of existing benchmarks. \textit{\#Scenario} indicates the number of categories, \textit{\#Instr.} reports the number of examples, \textit{Prompt}  specifies whether prompts are template-based (T), generated through an LLM (L) or taken from existing datasets (E), \textit{Layout} indicates whether instructions include bounding boxes.}
\label{tab:sota}
\end{table}

Recent advances in generative artificial intelligence have been largely driven by diffusion models, which now dominate text-to-image generation~\cite{DBLP:conf/nips/HoJA20,DBLP:conf/cvpr/RombachBLEO22,DBLP:journals/corr/abs-2211-01324,DBLP:journals/corr/abs-2112-10741,DBLP:conf/nips/SahariaCSLWDGLA22}. 
These models generate coherent and semantically faithful images from natural language prompts. 
Yet, many downstream applications, ranging from fashion image editing~\citep{DBLP:conf/iccv/BaldratiMCC0C23} to synthetic dataset generation~\citep{DBLP:conf/icpr/ParolariIB24}, require fine-grained control over image content. 
To support these applications, generated images must satisfy additional spatial or structural constraints, including layouts, sketches, or depth masks~\citep{DBLP:conf/cvpr/LiLWMYGLL23,DBLP:conf/iccv/XieLHLZ0S23,DBLP:conf/cvpr/ZhouLMZY24}. 
Among these control modalities, layout-guided generation has gained particular attention because of its simplicity and ease of use~\cite{DBLP:journals/tmlr/LianLYD24,DBLP:conf/iclr/GaniBN0W24}. 

Unlike standard text-to-image generation, evaluating layout-guided generation requires assessing two distinct yet complementary dimensions~\citep{DBLP:conf/wacv/GrimalBFT24}. 
A model must achieve semantic alignment, ensuring that the objects described in the prompt correctly appear in the image, while also maintaining spatial alignment, which measures whether these objects are positioned according to the prescribed layout. 

Collecting the annotations required for layout evaluation is costly and labor-intensive. 
It requires each textual prompt to be paired with a layout containing fine-grained bounding boxes linked to the noun phrases that describe the corresponding objects. 
Consequently, as summarized in Tab.~\ref{tab:sota}, most existing benchmarks do not include layout evaluation~\citep{DBLP:conf/nips/SahariaCSLWDGLA22,DBLP:conf/iclr/FengHFJANBWW23,DBLP:conf/iccv/Bakr0SKLE23,DBLP:conf/iccv/HuLKWOKS23,DBLP:conf/nips/HuangSXLL23,DBLP:conf/nips/WuYHRA24,DBLP:conf/iclr/GaniBN0W24,DBLP:journals/tmlr/LianLYD24}. 
Even when layout evaluation is included~\cite{DBLP:conf/cvpr/0001LYGWB22,DBLP:journals/corr/abs-2412-03859,DBLP:conf/nips/ChengMwLMWLY24,izzo20257bench}, the annotation cost often limits benchmark scale and leaves less room for broad coverage of specific generative skills~\citep{DBLP:conf/iclr/FengHFJANBWW23,izzo20257bench}. 
By contrast, benchmarks built on existing annotated datasets can scale to larger test sets, but they offer less control over the evaluation setup, making it harder to isolate and assess specific capabilities of layout-guided diffusion models~\citep{DBLP:journals/corr/abs-2412-03859,DBLP:conf/nips/ChengMwLMWLY24}. 

In this work, we introduce a \textit{closed-set benchmark} (\csb{}) for rigorous evaluation of layout-guided text-to-image models under controlled conditions. 
\csb{} defines precise evaluation scenarios that isolate specific generative skills, enabling targeted analysis of capabilities such as object or attribute binding and spatial reasoning, while offering different levels of instruction complexity. 
To complement this controlled setting, we introduce an \textit{open-set benchmark} (\osb{}) built from real-world data. 
\osb{} evaluates models using natural descriptions and real layouts, capturing the variability of human language and the structural diversity of real scenes. 
It therefore provides a more realistic measure of model generalization. 
Both these benchmarks do not require additional annotations.
Finally, we propose an evaluation protocol based on two complementary metrics: one measuring semantic alignment and the other measuring spatial alignment. 
These metrics offer a detailed characterization of model behavior, making it possible to distinguish failures in content generation from failures in layout adherence. 
We then combine them into a single \textit{unified score} that measures overall faithfulness to the input guidance. 
This metric enables comprehensive, interpretable assessment, supporting consistent ranking, direct comparison, and analysis of strengths and weaknesses in layout-guided image generation.

Using these benchmarks, we evaluate six state-of-the-art layout-guided diffusion models. 
Overall, we generate 319,086 images conditioned on the benchmark instructions. 
We assess these outputs with our evaluation protocol, obtaining semantic, spatial, and unified scores. 
This enables a clear ranking of current models, while the separate semantic and spatial scores provide a more interpretable view of their behavior. 
Fine-grained analyses across scenarios and varying prompt complexity further reveal the specific strengths and weaknesses of existing models, highlighting the compositional and spatial challenges that remain unresolved. 

\textbf{Contributions.}
\textit{(i)} We introduce a scalable closed-set benchmark to isolate key generative capabilities with controlled variations in prompt structure and layout. 
\textit{(ii)} We propose an open-set benchmark using real-world prompts and layouts to evaluate model generalization in natural settings.
\textit{(iii)} We develop a unified evaluation protocol combining text and layout alignment into a single score, enabling consistent model ranking.
\textit{(iv)} Using our benchmarks, we provide a model ranking based on the unified score.
\textit{(v)} We offer detailed performance breakdowns for text and layout alignment to enhance interpretability.
\textit{(vi)} We conduct fine-grained analyses across scenarios and prompt complexities to highlight the strengths and limitations of current models.

\begin{figure*}[ht]
\centering
\includegraphics[width=0.8\linewidth]{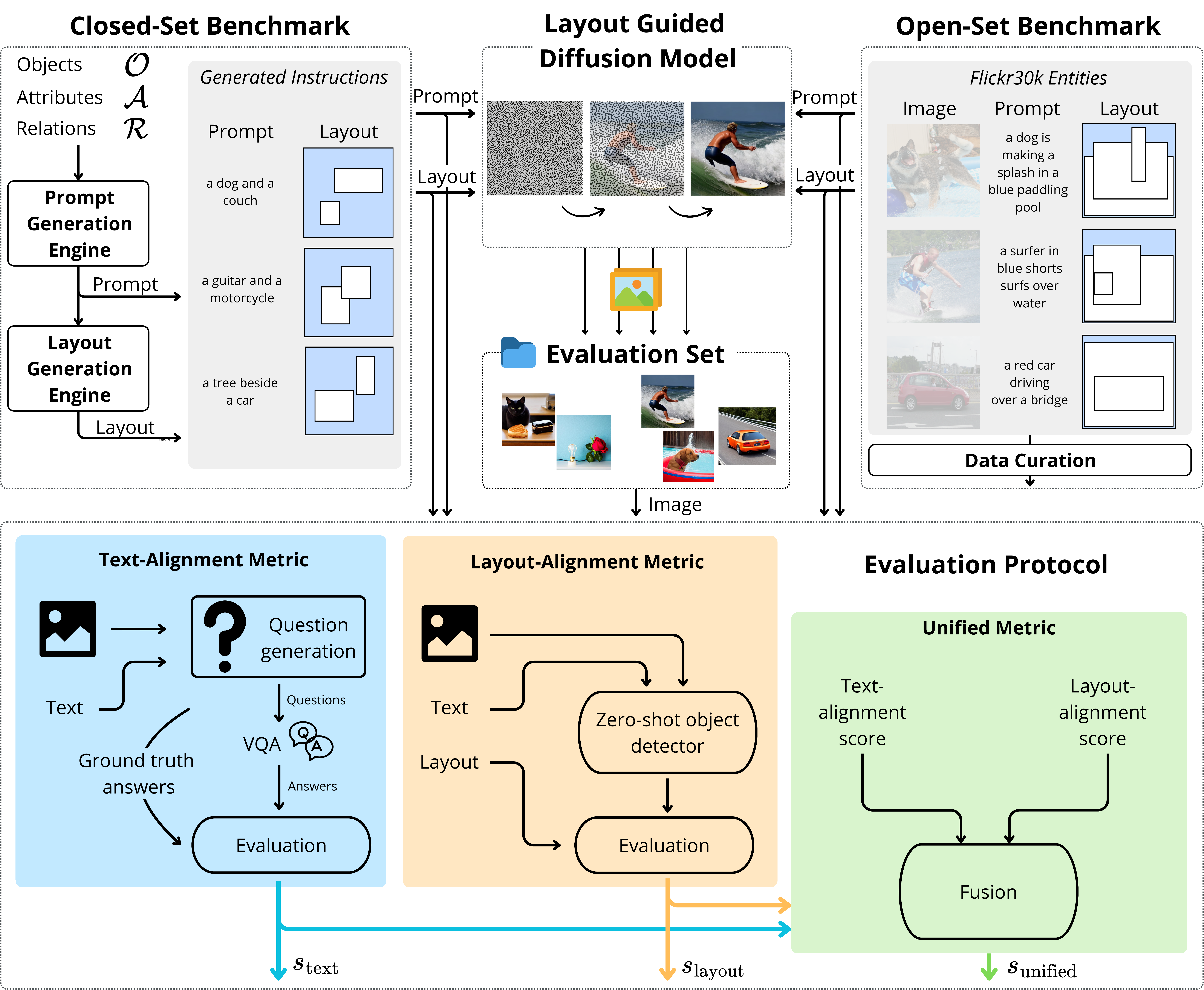}
\caption{
Overview of our evaluation framework. We construct the closed-set benchmark, automatically built by generating prompts and layouts, and the open-set benchmark derived from an existing dataset for visual grounding. A layout-guided diffusion model generates images, forming the evaluation set. The evaluation protocol assesses the images, providing text-alignment, layout-alignment and unified scores for both interpretability and consistent model ranking.
}
\label{fig:pipeline}
\end{figure*}

\section{Related Works}

Significant efforts have been devoted to evaluating the quality of images generated by text-to-image diffusion models (Tab.~\ref{tab:sota}).
Several works investigate prompt-image faithfulness by focusing on specific aspects such as color, attributes, counting, relations, or object composition. In these studies, prompts are either carefully crafted through templates~\citep{DBLP:conf/nips/SahariaCSLWDGLA22,DBLP:conf/iccv/HuLKWOKS23,DBLP:conf/nips/HuangSXLL23,DBLP:conf/iclr/FengHFJANBWW23}, 
generated by LLMs~\citep{DBLP:conf/nips/WuYHRA24,DBLP:conf/iclr/GaniBN0W24}, or a combination of both~\citep{DBLP:journals/tmlr/LianLYD24}.
In some cases, existing datasets are annotated or automatically processed to construct prompts from real data~\citep{DBLP:conf/iccv/Bakr0SKLE23,DBLP:conf/iclr/FengHFJANBWW23}.

However, with the emergence of layout-guided text-to-image models, most existing benchmarks lack two critical components: the input layout and a metric to quantify layout alignment.
As a result, evaluation practices for layout-guided generation remain heterogeneous and lack a unified protocol. For example, \citet{DBLP:conf/cvpr/LiLWMYGLL23} generate and evaluate 30k samples, whereas \citet{DBLP:conf/cvpr/PhungGH24} 5k. The COCO dataset~\citep{DBLP:conf/eccv/LinMBHPRDZ14} is sometime adopted to assess layout alignment~\citep{DBLP:journals/corr/abs-2412-03859,DBLP:conf/iclr/GaniBN0W24,DBLP:conf/nips/ChengMwLMWLY24}, despite not providing grounding annotations. Consequently, different metrics such as the YOLO score~\citep{DBLP:conf/iccv/LiWKTS21}, Average Precision~\citep{DBLP:conf/sigir/Robertson08}, or custom-designed measures~\cite{DBLP:conf/cvpr/ZhouLMZY24} are inconsistently employed.

More recently, Layout-Bench~\citep{DBLP:conf/cvpr/0001LYGWB22} introduced layout-based evaluation for inpainting but completely overlooks textual alignment.
HiCo~\citep{DBLP:conf/nips/ChengMwLMWLY24} proposes a Hierarchical Controllable Diffusion Model and evaluates it on HiCo-7K, a benchmark derived from GRIT-20M~\cite{DBLP:journals/corr/abs-2306-14824} that includes layout information.
Similarly, CreatiLayout~\citep{DBLP:journals/corr/abs-2412-03859} presents a Siamese Multimodal Diffusion Transformer and introduces LayoutSAM-Eval, which contains layout-image pairs with global captions and per-entity region descriptions.
Both do not offer fine-grained analysis of model's skills and are solely derived from existing datasets.

\section{Scalable Closed-set Benchmark}
\label{sec:closed-set}

We construct the closed-set benchmark using an automatic pipeline that generates the instruction set employed to probe generative models.
The pipeline consists of two main components. The Prompt Generation Engine (PGE) produces textual prompts by combining template-based rules and LLMs.
The Layout Generation Engine (LGE) then creates layouts through a constraint-based procedure, arranging bounding boxes in a realistic manner while controlling randomness according to the content of each prompt.
Despite its scalability, the pipeline preserves generality: the instruction set spans seven distinct scenarios and includes prompts with varying number of objects, thereby introducing structural and spatial complexity.
An overview of the pipeline is depicted in Fig.~\ref{fig:pipeline} (top-left).

\subsection{Prompt Generation Engine}

\begin{table*}
\centering
\begin{tabular}{ll}
\toprule
Scenario & Template \\
\midrule
Object binding & ``\textit{det}($\mathcal{o}_1$) $\mathcal{o}_1$, \textit{det}($\mathcal{o}_2$) $\mathcal{o}_2$, \textit{det}($\mathcal{o}_3$) $\mathcal{o}_3$  and \textit{det}($\mathcal{o}_4$) $\mathcal{o}_4$'' \\
Color binding & ``\textit{det}($\mathcal{c}_1$) $\mathcal{c}_1$ $\mathcal{o}_1$, \textit{det}($\mathcal{c}_2$) $\mathcal{c}_2$ $\mathcal{o}_2$, \textit{det}($\mathcal{c}_3$) $\mathcal{c}_3$ $\mathcal{o}_3$  and \textit{det}($\mathcal{c}_4$) $\mathcal{c}_4$ $\mathcal{o}_4$'' \\
Attribute binding & ``\textit{det}($\mathcal{a}_1$) $\mathcal{a}_1$ $\mathcal{o}_1$, \textit{det}($\mathcal{a}_2$) $\mathcal{a}_2$ $\mathcal{o}_2$, \textit{det}($\mathcal{a}_3$) $\mathcal{a}_3$ $\mathcal{o}_3$  and \textit{det}($\mathcal{a}_4$) $\mathcal{a}_4$ $\mathcal{o}_4$'' \\
Object relationship & ``\textit{det}($\mathcal{o}_1$) $\mathcal{r}_{12}$ \textit{det}($\mathcal{o}_2$) $\mathcal{o}_2$ and \textit{det}($\mathcal{o}_3$) $\mathcal{r}_{34}$ \textit{det}($\mathcal{o}_4$) $\mathcal{o}_4$'' \\
Small bboxes & ``\textit{det}($\mathcal{o}_1$) $\mathcal{o}_1$, \textit{det}($\mathcal{o}_2$) $\mathcal{o}_2$, \textit{det}($\mathcal{o}_3$) $\mathcal{o}_3$  and \textit{det}($\mathcal{o}_4$) $\mathcal{o}_4$'' \\
Overlapped bboxes & ``\textit{det}($\mathcal{o}_1$) $\mathcal{o}_1$, \textit{det}($\mathcal{o}_2$) $\mathcal{o}_2$, \textit{det}($\mathcal{o}_3$) $\mathcal{o}_3$  and \textit{det}($\mathcal{o}_4$) $\mathcal{o}_4$'' \\
\bottomrule
\end{tabular}
\caption{Templates used by the Prompt Generation Engine in different scenarios except complex compositions, shown with $N=4$ objects.}
\label{tab:templates}
\end{table*}

Textual prompts are generated using a combination of template-based rules and LLMs.
To ensure full control over the input and enable fair evaluation of outputs, we design seven scenarios with the goal to evaluate a precise generative capability.
Specifically, we investigate:
{
\renewcommand{\arraystretch}{1.5} 
\begin{tabular}{p{0.20\linewidth} p{0.7\linewidth}}
\makecell[lt]{Object\\binding} & The ability to generate elements described in the prompt; \\
\end{tabular}
\begin{tabular}{p{0.20\linewidth} p{0.7\linewidth}}
\makecell[lt]{Color\\binding} & The adherence of generated objects to color attributes; \\
\end{tabular}
\begin{tabular}{p{0.20\linewidth} p{0.7\linewidth}}
\makecell[lt]{Attribute\\binding} & The adherence to generic attributes like color, shape, material, appearance, and dimension; \\
\end{tabular}
\begin{tabular}{p{0.20\linewidth} p{0.7\linewidth}}
Object relationship & The ability to depict objects in relations, e.g. above, below, far from, to the left of; \\
\end{tabular}
\begin{tabular}{p{0.20\linewidth} p{0.7\linewidth}}
\makecell[lt]{Small\\bboxes} & The ability to depict objects whose area is between 3 and 10\% of the image area; \\
\end{tabular}
\begin{tabular}{p{0.20\linewidth} p{0.7\linewidth}}
\makecell[lt]{Overlapped\\bboxes} & The ability to depict objects that overlaps in terms of layout; \\
\end{tabular}
\begin{tabular}{p{0.20\linewidth} p{0.7\linewidth}}
\makecell[lt]{Complex\\compositions} & The ability to depict a scene described by arbitrary mixing of previous scenarios and layout. \\
\end{tabular}
}

\noindent For each scenario except complex compositions, we design a template that strictly constrains text structure, object-attribute combinations, and inter-object relations, thereby isolating the intended capability under evaluation.  
The templates are described with a formalism inspired by the disentangle representation theory~\citep{DBLP:conf/iccv/TragerPZABS23}, and are reported in Tab.~\ref{tab:templates}. 
Each prompt describes up to four objects. An example of a template for the case where \textit{N}$=2$ objects is:
\begin{equation}
    \textit{t} = \text{``\textit{det}($\mathcal{o}_1$,$\mathcal{a}_1$) $\mathcal{a}_1$ $\mathcal{o}_1$ \textit{rel}(\textit{and}, $\mathcal{r}_{12}$) \textit{det}($\mathcal{o}_2$,$\mathcal{a}_2$) $\mathcal{a}_2$ $\mathcal{o}_2$''}
\end{equation}
where, $\mathcal{o}_i \in \mathcal{O}$ the set of objects, $\mathcal{a}_i \in \mathcal{A}$ the set of attributes, \textit{det}($\mathcal{o}_i$,$\mathcal{a}_i$) is a determinant (e.g. a, an) that depends on the object or attribute if present, and \textit{rel}(\textit{and}, $\mathcal{r}_{ij}$) is the coordinating conjunction \textit{and} or a relation $\mathcal{r}_{ij} \in \mathcal{R}$--the set of relations--if present.
To generate actual prompts, templates are instantiated by randomly picking objects, attributes and relations from their respective sets.

Complex compositions scenario involves mixing all the previously isolated challenges into a single setting. Instead of relying on rigid templates we employ an LLM, well suited for fluent and contextually rich sentences.
The prompts are generated by an LLM conditioned on the same sets of objects, attributes, and relations.
Specifically, we employ few-shot learning~\citep{DBLP:conf/nips/BrownMRSKDNSSAA20} and provide instructions accompanied by example outputs. 
The LLM is given $\mathcal{O}$, $\mathcal{A}$, and $\mathcal{R}$ and instructed to combine them freely into coherent descriptions, while also guided to mention a specified number of objects (1-4), to enable fine-grained analysis.
We refer the reader to the Supplementary Material for details on the sets $\mathcal{O}, \mathcal{A}, \mathcal{R}$ and the prompt used to instruct the LLM.

\subsection{Layout Generation Engine}

To collect the large number of bounding boxes associated with the prompts without relying on costly manual annotation, we designed a constraint-based Layout Generation Engine (LGE). 
It produces a reasonable layout for any given prompt. 
Specifically, we first obtain the set $\{o_{j,i}\}_{j=1}^N$ of objects described in the $i$-th prompt. In case this set is not provided by the Prompt Generation Engine (e.g. when prompts are LLM-generated), we extract the relevant objects automatically with a natural language parser such as spaCy~\citep{spacy2020}. Then, for each object $\mathcal{o}_{j,i}$, the LGE produces a bounding box: $b_{j,i} = [x^\text{min}_j,y^\text{min}_j,x^\text{max}_j,y^\text{max}_j]$ with $0 \leq x^\text{min}_j < x^\text{max}_j \leq 1$ and $0 \leq y^\text{min}_j < y^\text{max}_j \leq 1$.

Bounding box coordinates are chosen according to different constraints. 
For single-object prompts, the box is placed randomly within the image boundaries. For multi-object prompts, placement strategies vary: if spatial relations are involved (e.g., above, far from, to the left of), boxes are positioned accordingly.
The first box is sampled randomly, leaving sufficient space for subsequent boxes to satisfy the relation; invalid placements trigger retries until a valid configuration is obtained. 
In case of overlapped bboxes scenario, subsequent sampled boxes are forced to overlap with at least one previously placed box. 
For prompts with multiple objects but no explicit relations, bounding boxes are placed randomly while enforcing non-overlap through rejection sampling, while complex compositions scenario does not follow any constraint.

\section{Open-set Benchmark}
\label{sec:open-set}

While the closed-set benchmark evaluates generative models under controlled conditions--using structured prompts and carefully designed bounding boxes, the open-set benchmark is intended to assess them in more natural settings.
However, instead of relying on costly manual annotation of prompts and bounding boxes, we re-frame the purpose of Flickr30k Entities~\citep{DBLP:journals/ijcv/PlummerWCCHL17}, a widely used dataset in fine-grained visual-language tasks such as Visual Grounding~\citep{DBLP:conf/iccv/DengYCZL21,DBLP:conf/bmvc/RigoniPSSB23}. Flickr30k Entities provides real-world images paired with multiple human-annotated captions, each associated with bounding boxes localizing the mentioned objects. 
We employ these realistic prompts and unconstrained layouts to construct our open-set benchmark: this process is visually depicted in Fig.~\ref{fig:pipeline} (top-right).

Specifically, we construct our open-set benchmark (\osb{}) on the Flickr30k Entities test set~\citep{DBLP:journals/ijcv/PlummerWCCHL17}, to avoid potential data leakage with the training sets of layout-guided text-to-image models. 
The test set comprises 4,969 captions which we downsample to approximately two-thirds of its original size, resulting in 3,319 prompts. 
We downsample the test set to avoid excessively large datasets, which would make evaluation impractical given the slow inference speed of current models. As later described in the evaluation setup, we generate 8 images per example with different random seeds to ensure robustness in the evaluation. This greatly increases computation time: for instance, BoxDiff~\citep{DBLP:conf/iccv/XieLHLZ0S23} requires over 13 GPU-days to process the full test set, while the downsampled version reduces this to 7.81 GPU-days.
Sampling was performed to preserve the original distribution of objects per sentence.

Formally, the instruction set constituting the open-set benchmark is defined as a collection of prompt-object pairs: \osb{} $ = \{(T_i, O_i)\}_{i=1}^N$, where each element is derived from the Flickr30k test set. For each sample, $T_i$ denotes the textual description associated with the image (note that the image itself is discarded), and $O_i = \{(o_1, b_1), \ldots, (o_M, b_M)\}$ represents the set of objects mentioned in the caption. Each object $o_j$ is described by a noun phrase (e.g., a small red umbrella) and paired with its corresponding spatial location $b_j$.
The final benchmark includes prompts describing different real-life situations, offering a realistic and representative range for evaluation.

\section{Evaluation Protocol}
\label{sec:eval}

To evaluate the alignment between generated images and their input prompts and layouts, we propose two complementary metrics that measure semantic alignment and spatial fidelity: the text-alignment score $s_\text{text}$ and the layout-alignment score $s_\text{layout}$. 
Furthermore, we introduce another metric that enables precise ranking of models by combining both dimensions in one single score: the unified score $s_\text{unified}$. 
An overview is given in Fig.~\ref{fig:pipeline} (bottom).

\subsection{Text-Alignment Score}

We define the text-alignment score as the TIFA score~\citep{DBLP:conf/iccv/HuLKWOKS23}, a widely adopted measure of semantic consistency in text-to-image generation. 
TIFA quantifies alignment as the proportion of correct responses provided by a Vision Question Answering (VQA) model, such as BLIP-2~\citep{DBLP:conf/icml/0008LSH23}, when analyzing the generated images. 
To perform the evaluation, a set of questions and corresponding answers are automatically derived from the input prompt using a Large Language Model, namely Llama2~\citep{DBLP:journals/corr/abs-2307-09288}, ensuring independence from the image generation process. 
After a filtering process operated by a Question Answering model such as UnifiedQA~\citep{DBLP:conf/emnlp/KhashabiMKSTCH20}, the VQA model is then queried on the generated image, and its responses are compared against the expected answers to compute the final score. 
Formally, let $\mathcal{X}=\{(T_i, I_i)\}_{i=1}^N$ be the set of input text prompts and generated images. For each example $i$, a set of questions, expected answers and actual answer is obtained: $\{Q_{j,i}, A_{j,i}, A_{j,i}^{\text{VQA}}\}_{j=i}^{M_i}$. The text alignment score for example $i$ is defined as the average accuracy of VQA answers over expected answers. The overall score is then obtained as the mean across all examples:
\begin{equation}
s_{\text{text}}(i) = \frac{1}{M_i} \sum_{j=1}^{M_i} \mathbf{1} [ A^{\text{VQA}}_{j,i} = A_{j,i}]
\qquad
s_{\text{text}} = \frac{1}{N}\sum_{i=1}^{N} s_{\text{text}}(i).
\end{equation}

To better understand model behavior, we extend the base score $s_{\text{text}}$ by conditioning the evaluation on two axes: the scenario and the number of objects in the prompt. Analyzing the score along one or the other enhances interpretability in the evaluation process.
Let $\mathcal{C}$ be the set of scenarios (e.g., object binding, color binding, etc), and $\mathcal{D}=\{1,2,3,4\}$ the number of objects in the prompt.
Let $\sigma(i)\in\mathcal{C}$ be the scenario of example $i$, and $\nu(i)\in\mathcal{D}$ its number of objects.
We obtain the set of indices $\mathcal{Z}_{c,d} = \{i\in\{1,\dots,N\} : \sigma(i)=c,\ \nu(i) = d\}$ and overload the definition of $s_{\text{text}}$:
\begin{equation}
\begin{gathered}
s_{\text{text}}^{c,d} = \frac{1}{|\mathcal{Z}_{c,d}|}\sum_{i\in \mathcal{Z}_{c,d}} s_{\text{text}}(i),
\qquad
s_{\text{text}}^{d} = \frac{1}{|\mathcal{Z}_d|}\sum_{i \in \mathcal{Z}_{c,d} \forall c} s_{\text{text}}^{c,d},
\\
s_{\text{text}}^{c} = \frac{1}{|\mathcal{Z}_c|}\sum_{i \in \mathcal{Z}_{c,d} \forall d} s_{\text{text}}^{c,d}.
\end{gathered}
\end{equation}
For example, $s^2_\text{text}$ is the text-alignment score on all examples with exactly two objects in the prompt from any scenario, while $s^\text{object binding}_\text{text}$ is the text-alignment score on all examples from object binding scenario and any number of objects in the prompt.

\subsection{Layout-Alignment Score}

We define the layout-alignment score $s_\text{layout} \in [0, 1]$ as the Area Under Curve (AUC) of accuracy@k values computed over a range of Intersection over Union (IoU) thresholds. The $s_\text{layout}$ score captures the spatial accuracy of objects' placement within the generated images.

Specifically, let $\mathcal{X}=\{(T_i, I_i, O_i)\}_{i=1}^N$ be the set of examples to evaluate. $T_i$ is the input text prompt, $I_i$ is the generated image and $O_i = \{(o_1, b_1), \ldots, (o_M, b_M)\}$ represents the set of objects mentioned in the prompt along with their spatial locations. 
Using a zero-shot object detector such as OWLv2~\citep{minderer2023scaling}, we obtain a set of $K$ detections for each $o_{j,i}, j \in M$ accompanied by confidence scores: $D_{j,i} = \{(d_k, c_k)\}_{k=1}^K$.
From $D_{j,i}$ we select the bounding box with the higher confidence score $\hat{d}_{j,i} = \arg\max_{(d_k, c_k) \in D_{j,i}} c_k$.
We then compute the IoU between ground truth and predicted locations of the object $j$: $\text{IoU}_{j,i} = \text{IoU}(b_{j,i}, \hat{d}_{j,i})$.
Subsequently, these IoU values are thresholded at multiple levels $k \in \{0, 0.1, \ldots, 1\}$ to calculate $\text{Acc@}k_{j,i} = \frac{1}{M} \sum_{j=1}^{M} \mathbf{1}[\text{IoU}_{j,i} \geq k]$.
The layout-alignment score for the single example $i$ is the area under the resulting $\text{Acc@}k$ curve, and the overall score is averaged across all examples:
\begin{equation}
s_{\text{layout}}(i) = \frac{1}{M} \sum_{j=1}^{M} \text{AUC}( \text{Acc@}k_{j,i}),
\quad
s_{\text{layout}} = \frac{1}{N} \sum_{i=1}^N s_{\text{layout}} (i).
\end{equation}
Similarly to the text-alignment score, we extend the formulation to evaluate along different axis: scenario $s_{\text{layout}}^{c}(i)$, number of object $s_{\text{layout}}^{d}(i)$ and both $s_{\text{layout}}^{c,d}(i)$.

\subsection{Unified Score}

Ranking the performance of layout-guided text-to-image models poses a complex challenge, as it requires capturing both semantic alignment with the input text and spatial fidelity to the given layout. While text and layout alignment enable fine-grained interpretability of the generation capabilities, a unified metric that accounts for both aspects is necessary to compare and rank different models.
This metric must provide a balanced assessment across both dimensions: it should reflect strong performance only when a model is consistent in both text and layout, and it should penalize cases of disagreement, such as when an image is semantically correct but spatially inconsistent, or vice versa. In this way, the overall score captures the true alignment quality.

Following previous literature in multi-objective evaluation settings, such as the F1 score in information retrieval~\citep{DBLP:journals/jasis/Blair79}, we defined the unified score $s_\text{unified}$ as the harmonic mean between $s_\text{text}$ and $s_\text{layout}$:
\begin{equation}
s_\text{unified} = H(s_\text{text}, s_\text{layout}) = \frac{ 2 \cdot s_\text{text} \cdot s_\text{layout} }{ s_\text{text} + s_\text{layout} }.
\end{equation}
The harmonic mean penalizes imbalances between the two components--being proportional to the product over the sum--ensuring that strong performance requires both textual and spatial consistency. 
Similarly to text and layout alignment scores, we extend the analysis of $s_\text{unified}$ along scenarios, number of objects and both: $s_{\text{unified}}^{c}$, $s_{\text{unified}}^{d}$ and $s_{\text{unified}}^{c,d}$.

\section{Experimental Evaluation}

\subsection{Setup}
\paragraph{\csb{}}
Following the pipeline described in Sec.~\ref{sec:closed-set}, we generated the closed-set benchmark (\csb{}), resulting in 3,328 instructions.
Thanks to the scalability of our pipeline, \csb{} ensures a balanced number of prompts per scenario, maintains a fair distribution of object occurrences, and provides comprehensive coverage of bounding box sizes without requiring costly manual annotations.
From a technical standpoint, each sample in \csb{} consists of a textual prompt paired with a set of bounding boxes, one for each object described. 
Furthermore, for every object, the benchmark also provides a noun phrase that specifies the entity together with its attributes and qualifiers.
Prompts for the complex compositions scenario are generated by ChatGPT 4o~\citep{hurst2024gpt}.

\smallskip
\noindent\textbf{Models Under Test}\quad
We evaluate 6 popular layout-guided text-to-image diffusion models on our closed and open-set benchmarks. Models are all open source and have been accurately chosen to explore a wide range of methods and techniques. In particular, we test GLIGEN~\citep{DBLP:conf/cvpr/LiLWMYGLL23} and MIGC~\citep{DBLP:conf/cvpr/ZhouLMZY24}, which are trained with layout information, and three training-free approaches: Attention Refocusing~\citep{DBLP:conf/cvpr/PhungGH24}, BoxDiff~\citep{DBLP:conf/iccv/XieLHLZ0S23}, and Cross-Attention Guidance~\citep{DBLP:conf/wacv/ChenLV24}. The first two are built on top of GLIGEN, the last one uses Stable Diffusion as the underlying model. Finally, we include Stable Diffusion~\citep{DBLP:conf/cvpr/RombachBLEO22} in the analysis for a comparison in terms of textual alignment.

\smallskip
\noindent\textbf{Evaluation Setting}\quad
Following previous works~\citep{DBLP:conf/wacv/GrimalBFT24,izzo20257bench}, we generate 8 images for each instruction in the benchmark, varying the seed from $1$ to $8$ to ensure robustness against sampling variability and to obtain a reliable estimate of model performance. 
Across all models under test, the generation produced a total of 159,744 and 159,312 images for \csb{} and \osb{}, respectively. Each image was generated with a resolution of $512 \times 512$.
We evaluate the generated images using the evaluation protocol described in Section~\ref{sec:eval}. 
We use pre-trained weights for TIFA\footnote{\href{https://github.com/Yushi-Hu/tifa}{https://github.com/Yushi-Hu/tifa}} and OWLv2\footnote{\href{https://huggingface.co/docs/transformers/model_doc/owlv2}{https://huggingface.co/docs/transformers/model\_doc/owlv2}}.

\smallskip
\noindent\textbf{Human Validation of Pipeline Components}\quad
To assess the reliability and sensitivity of key components in our evaluation framework, we performed two user studies.
First, we measured the agreement on semantic alignment between TIFA~\citep{DBLP:conf/iccv/HuLKWOKS23} and human judgments. We manually evaluated a subset of 640 generated images from different models, checking the adherence to the given layout. We compared our judgments to TIFA scores and found good overall consistency, with minor discrepancies in color-binding cases.
Second, we assessed the reliability of the object detector~\citep{DBLP:journals/corr/abs-2205-06230} employed for layout-alignment evaluation. A subset of 448 generated images was manually annotated with bounding boxes to evaluate layout adherence, and detector predictions were compared with human annotations by measuring their spatial overlap. The evaluation showed closely aligned behaviors, with humans exhibiting slightly higher localization precision. Further details are reported in the Supplementary Material.

\subsection{Results}

\begin{table*}
\centering
\begin{tabular}{l|cc|cc}
\toprule
& \multicolumn{2}{c}{Closed-set} & \multicolumn{2}{|c}{Open-set} \\
Model & $s_\text{unified}$ & $\Delta\%$ & $s_\text{unified}$ & $\Delta\%$ \\
\midrule
MIGC~\citep{DBLP:conf/cvpr/ZhouLMZY24} & \textbf{0.7082} & \footnotesize\textcolor{gray}{+0.0} & \textbf{0.7548} & \footnotesize\textcolor{gray}{+0.0} \\
BoxDiff~\citep{DBLP:conf/iccv/XieLHLZ0S23} & \underline{0.6537} & \footnotesize\textcolor{red}{$-8.4$} & 0.7410 & \footnotesize\textcolor{red}{$-1.8$} \\
GLIGEN~\citep{DBLP:conf/cvpr/LiLWMYGLL23} & 0.6143 & \footnotesize\textcolor{red}{$-13.3$} & \underline{0.7517} & \footnotesize\textcolor{red}{$-0.1$} \\
Attention Refocusing~\citep{DBLP:conf/cvpr/PhungGH24} & 0.6070 & \footnotesize\textcolor{red}{$-14.3$} & 0.7305 & \footnotesize\textcolor{red}{$-3.2$} \\
Cross-Attention Guidance~\citep{DBLP:conf/wacv/ChenLV24} & 0.3747 & \footnotesize\textcolor{red}{$-47.1$} & 0.5370 & \footnotesize\textcolor{red}{$-28.9$} \\
Stable Diffusion\textsuperscript{*}~\citep{DBLP:conf/cvpr/RombachBLEO22} & 0.2522 & \footnotesize\textcolor{red}{$-64.4$} & 0.4505 & \footnotesize\textcolor{red}{$-40.3$} \\ 
\bottomrule
\end{tabular}
\caption{Ranking for the six models under test on closed (\csb{}) and open set (\osb{}) benchmarks. \textsuperscript{*}Stable Diffusion is reported as baseline since it does not have layout capabilities. \textbf{Bold} = best model, \underline{Underline} = second best. $s_\text{unified}$ represents our unified score and $\Delta\%$ is the performance delta between model's score and top performer.}
\label{tab:model-ranking}
\end{table*}

\begin{figure*}
\begin{subfigure}[b]{0.48\linewidth}
    \includegraphics[width=\linewidth]{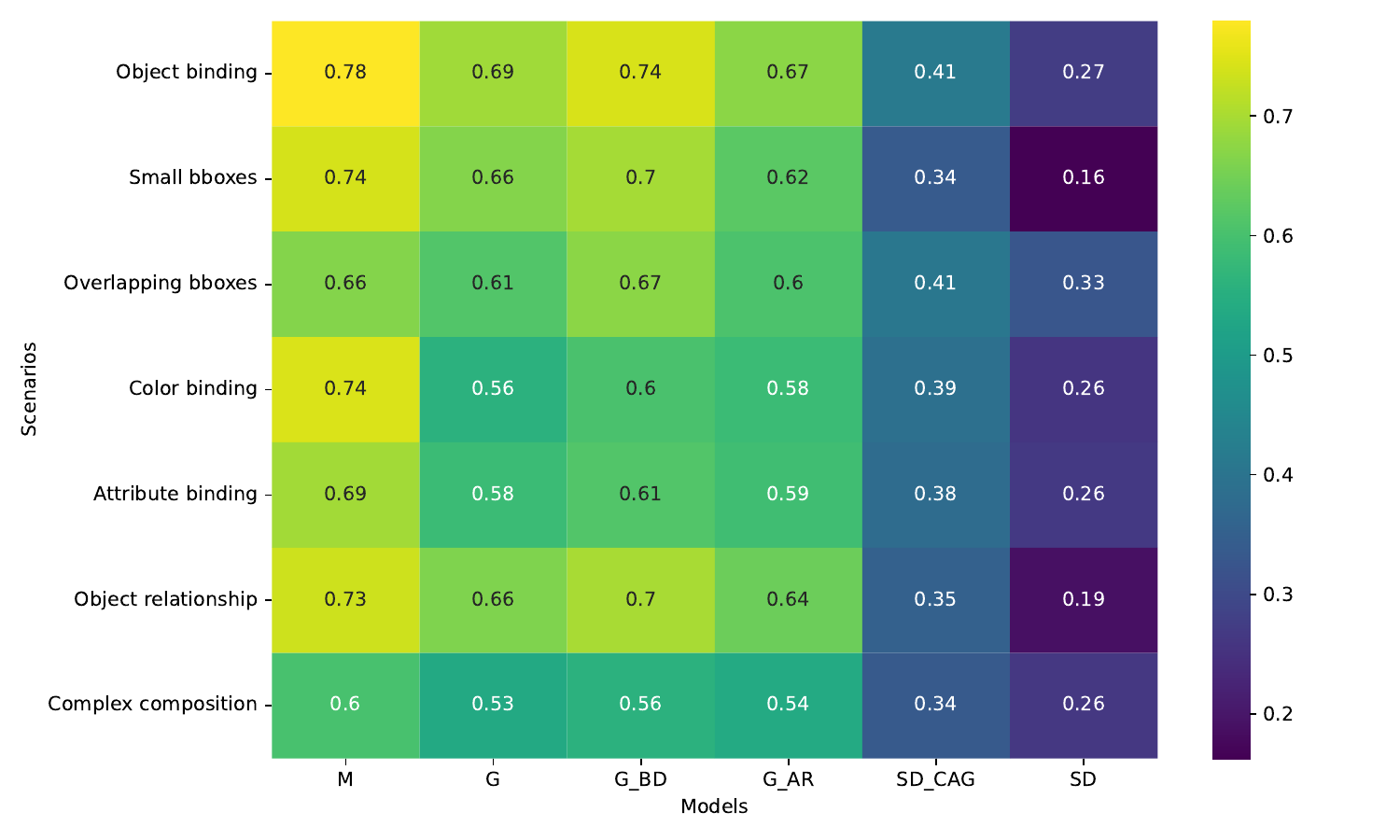}
    \caption{Performance by scenario measured with the unified score.}
    \label{fig:res-unified-scenario}
\end{subfigure}
\hfill
\begin{subfigure}[b]{0.48\linewidth}
    \includegraphics[width=\linewidth]{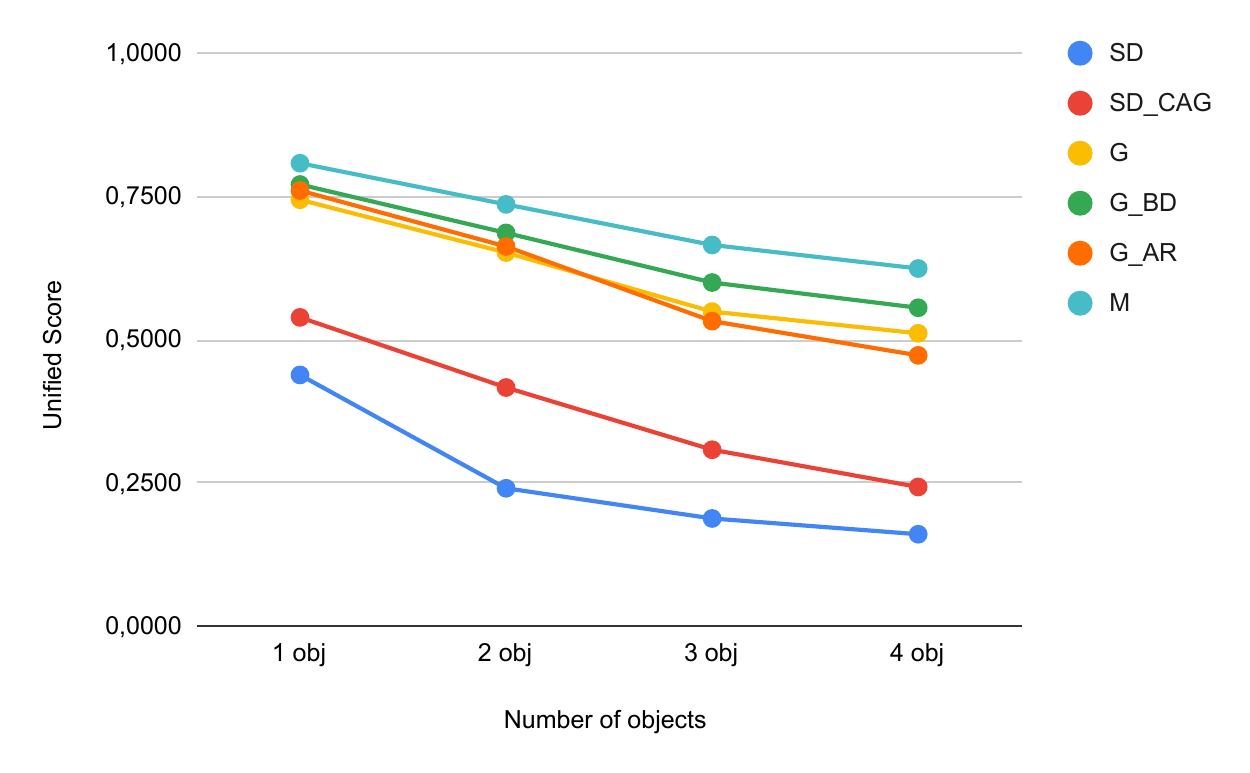}
    \caption{Performance by number of objects in prompt.}
    \label{fig:res-unified-nobj}
\end{subfigure}
\caption{Performance by scenario and object count on \csb{}, measured by the unified score for the six models under test. SD = Stable Diffusion~\citep{DBLP:conf/cvpr/RombachBLEO22}, SD\_CAG = Cross-Attention Guidance~\citep{DBLP:conf/wacv/ChenLV24}, G = GLIGEN~\cite{DBLP:conf/cvpr/LiLWMYGLL23}, G\_BD = BoxDiff~\citep{DBLP:conf/iccv/XieLHLZ0S23}, G\_AR = Attention Refocusing~\citep{DBLP:conf/cvpr/PhungGH24}, M = MIGC~\citep{DBLP:conf/cvpr/ZhouLMZY24}.}
\label{fig:res-unified}
\end{figure*}

We present the ranking on both closed and open set benchmarks according to $s_\text{unified}$ in Tab.~\ref{tab:model-ranking}.
From our evaluation MIGC achieves a unified score of 0.7082 and 0.7548 on \csb{} and \osb{} respectively, signaling its robustness in both semantic and spatial aspects of generation. 
GLIGEN-based models also obtain decent performance (0.6070-0.6537, 0.7305-0.7517), showing the impact of pre-training with layout information. Cross-Attention Guidance--which is based on Stable Diffusion--obtains a unified score of 0.3747 and 0.5370: a considerable drop in performance with respect to the top-performing model.

\smallskip
\noindent
\textbf{Unified Score Analysis}\quad 
We further analyze the behavior of $s_\text{unified}$ in terms of sensitivity and ranking stability. 
When progressively perturbing the benchmark instructions while keeping the generated images fixed, the unified score and its semantic and spatial components decrease smoothly and monotonically, indicating predictable sensitivity to increasing mismatches. 
Moreover, rankings induced by $s_\text{unified}$ remain highly consistent with those obtained from linear combinations of $s_{\text{text}}$ and $s_{\text{layout}}$ (average Kendall's $\tau$: 0.9152, Spearman's $\rho$: 0.9584), showing that the ranking is not an artifact of the chosen aggregation. 
Additional details are provided in the Supplementary Material. 

\subsection{Closed-set Results Breakdown}

\smallskip
\noindent\textbf{Breakdown by Scenario and Object Count}\quad
To further interpret the ranking and provide a more detailed view of model performance, we also report results broken down by scenario and by the number of objects in the prompt.
Fig.~\ref{fig:res-unified-scenario} confirms the overall ranking, with MIGC achieving the highest scores across almost all scenarios. As expected~\citep{DBLP:conf/wacv/GrimalBFT24}, the complex compositions scenario proves more challenging than the others, with smaller performance gaps among models.
Fig.~\ref{fig:res-unified-nobj} additionally illustrates how performance decreases as the number of objects in the prompt increases. All models exhibit a decline in accuracy when handling more objects. Interestingly, aside from Stable Diffusion--which does not support layout input--models show a similar degradation pattern, suggesting comparable sensitivity to object count.

\begin{figure*}[ht]
\begin{subfigure}[b]{0.48\linewidth}
    \includegraphics[width=0.9\linewidth]{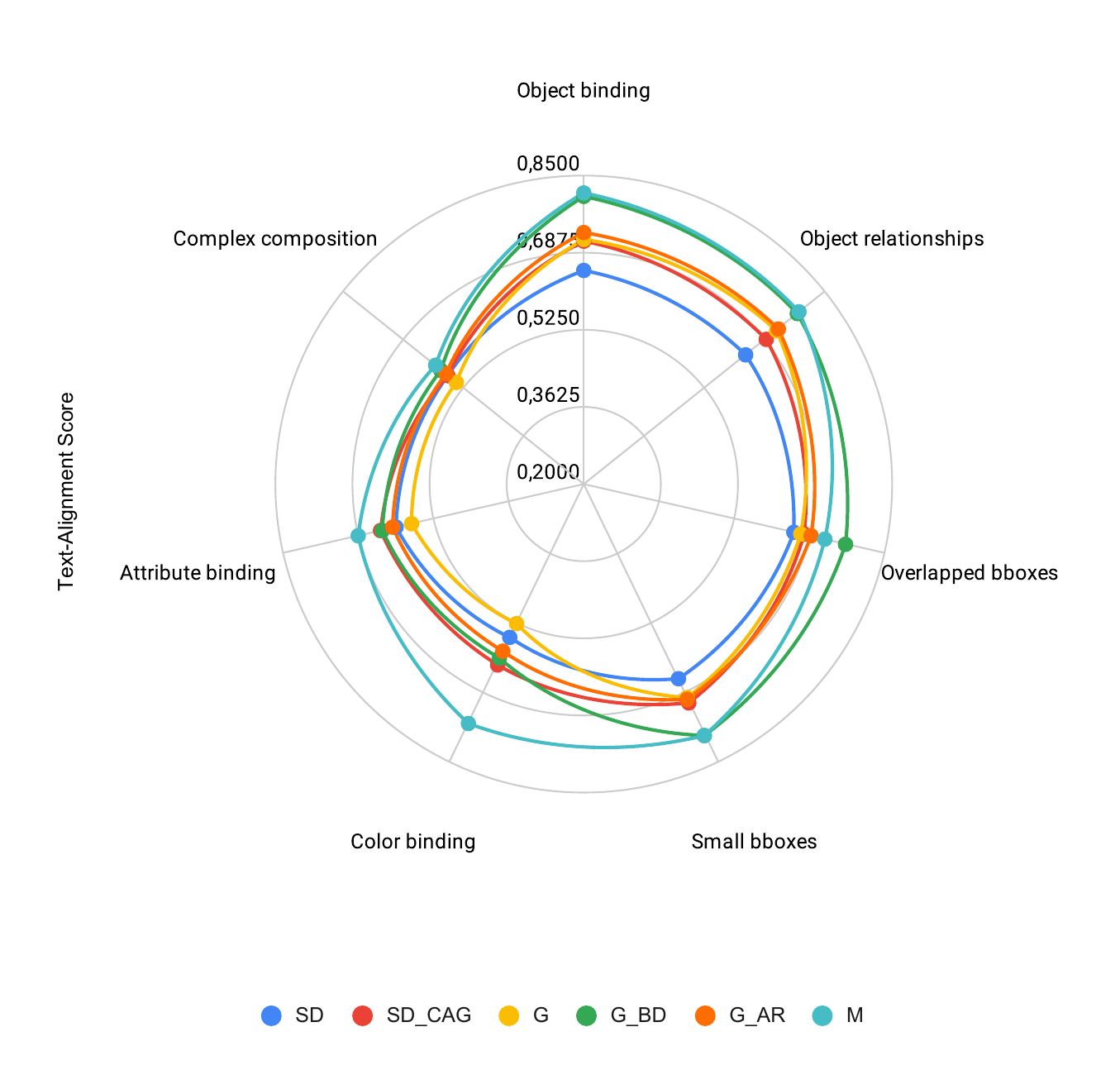}
\end{subfigure}
\hfill
\begin{subfigure}[b]{0.48\linewidth}
    \includegraphics[width=0.9\linewidth]{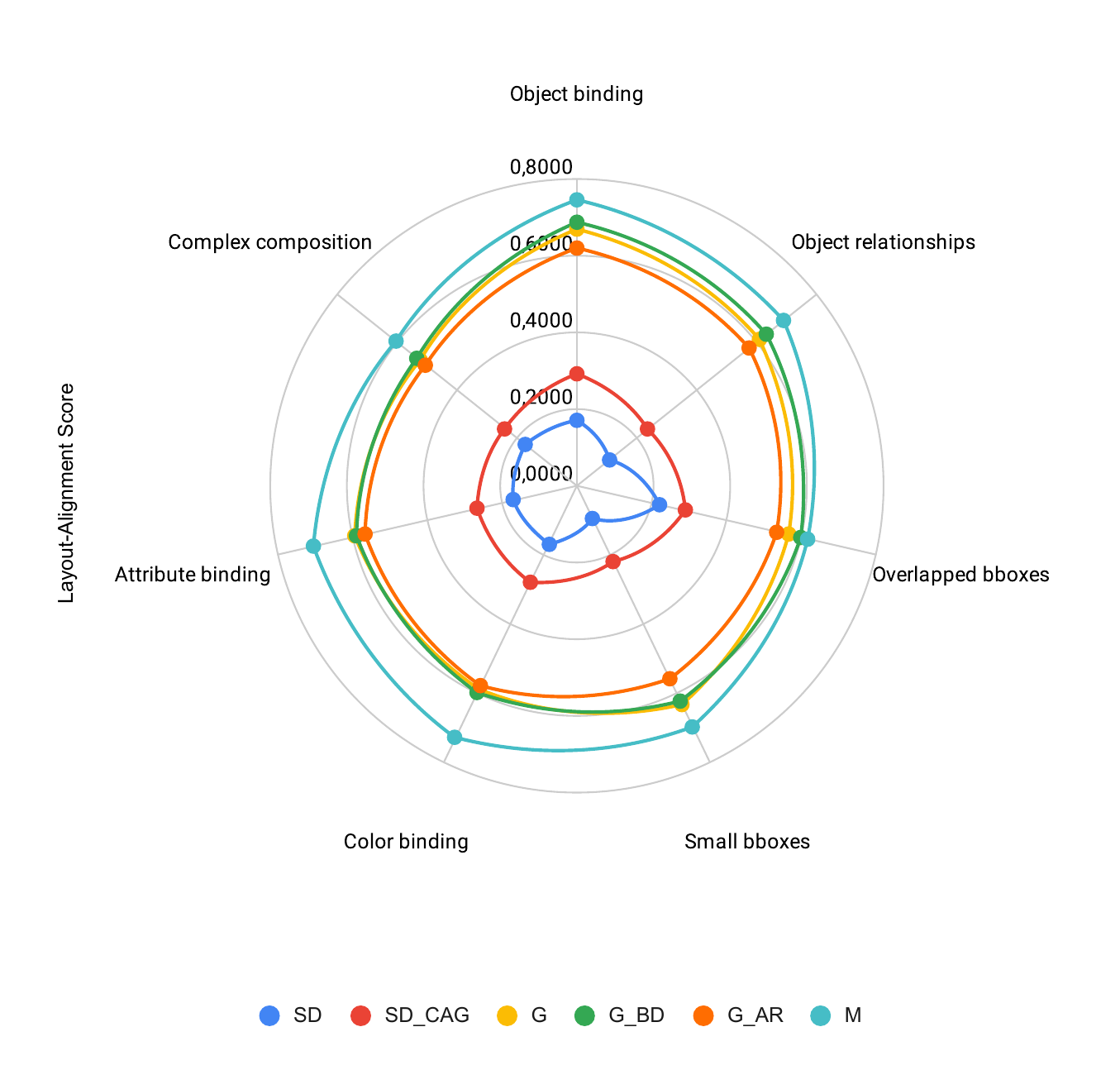}
\end{subfigure}
\begin{subfigure}[b]{0.48\linewidth}
    \includegraphics[width=\linewidth]{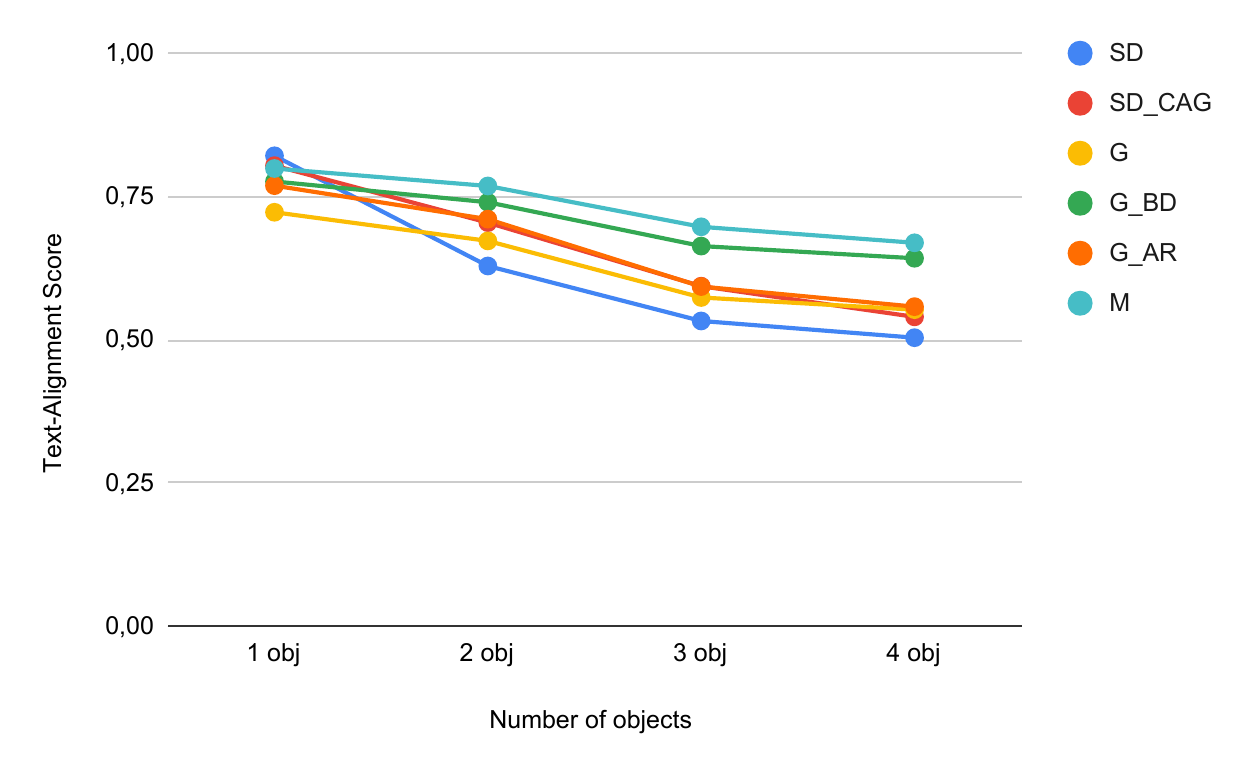}
\end{subfigure}
\hfill
\begin{subfigure}[b]{0.48\linewidth}
    \includegraphics[width=\linewidth]{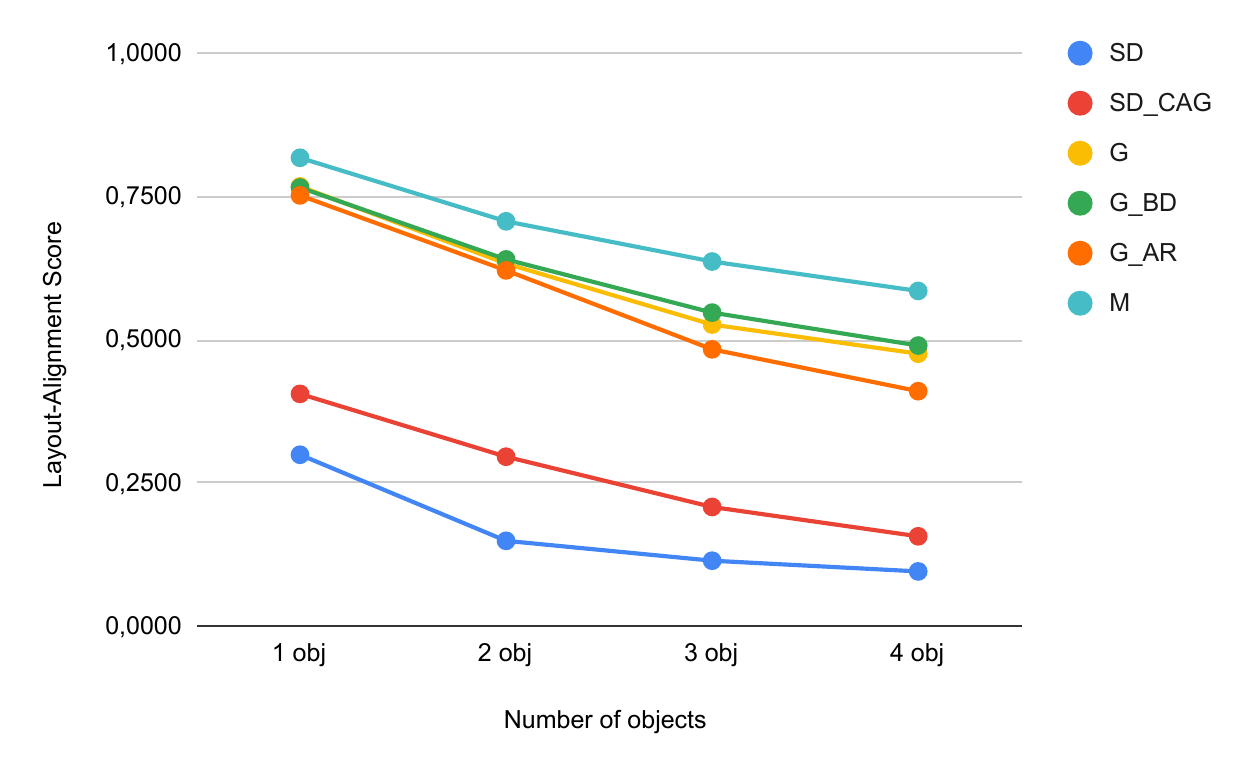}
\end{subfigure}
\caption{Performance breakdown by scenario (top) and number of objects in the prompt (bottom) on \csb{}, measured by text-alignment and layout-alignment score (resp. left, right). SD = Stable Diffusion~\citep{DBLP:conf/cvpr/RombachBLEO22}, SD\_CAG = Cross-Attention Guidance~\citep{DBLP:conf/wacv/ChenLV24}, G = GLIGEN~\cite{DBLP:conf/cvpr/LiLWMYGLL23}, G\_BD = BoxDiff~\citep{DBLP:conf/iccv/XieLHLZ0S23}, G\_AR = Attention Refocusing~\citep{DBLP:conf/cvpr/PhungGH24}, M = MIGC~\citep{DBLP:conf/cvpr/ZhouLMZY24}.}
\label{fig:res-breakdown-scenario-nobj}
\end{figure*}

\smallskip
\noindent\textbf{Breakdown on Text and Layout Alignment}\quad
Fig.~\ref{fig:res-breakdown-scenario-nobj} shows text-alignment and layout-alignment scores, revealing that the primary source of errors stems from layout alignment.
MIGC demonstrates strong performance in layout adherence across all scenarios. 
Interestingly, while Cross-Attention Guidance performs well on $s_\text{text}$, its performance drops sharply for $s_\text{layout}$. 
This drop explains why, as shown in the previous section, our unified metric $s_\text{unified}$ ranks Cross-Attention Guidance fifth out of six models, effectively penalizing good text alignment when accompanied by poor layout fidelity.
This trend is consistent also when analyzing the results by number of objects, confirming that layout-aware models maintain an advantage even at higher complexity.
While text and layout alignment alone are not sufficient to evaluate overall performance, they provide valuable interpretability into the sources of errors.

\subsection{Open-set Results Breakdown}

\begin{figure}
    \centering
    \includegraphics[width=\linewidth]{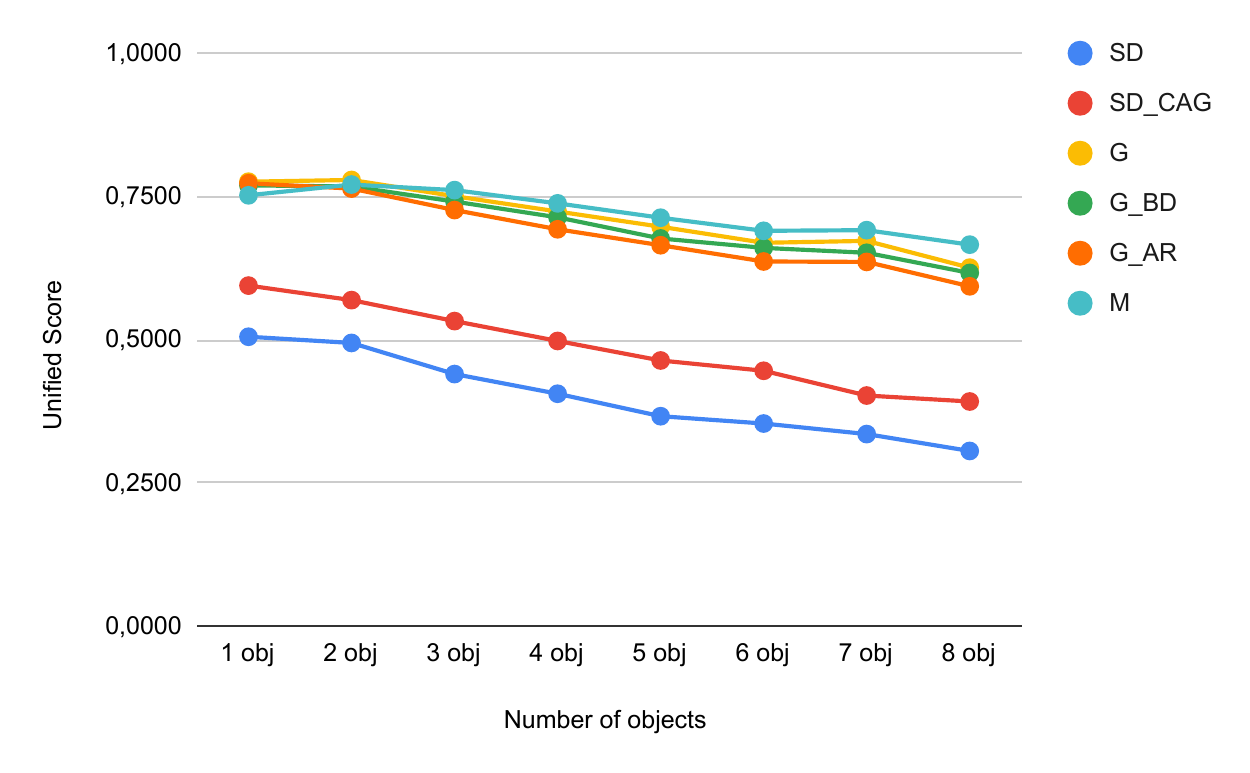}
    \caption{Performance breakdown by object count on \osb{}, measured with the unified score.}
    \label{fig:open-set-nobj-unified}
\end{figure}

We further analyze the results obtained on \osb{} and provide the interpretation of global ranking by showing results broken down by object count. 
We recall that the open-set benchmark includes prompts describing different numbers of objects but, due to its open nature, does not define specific scenarios. 
As shown in Fig.~\ref{fig:open-set-nobj-unified}, model performance decreases as the number of objects increases. 
As in closed-set case, all models exhibit a comparable degradation pattern, indicating consistent sensitivity to object count.

\section{Conclusion}

We present a framework for evaluating layout-guided text-to-image generative models. By introducing a scalable closed-set benchmark, an open-set benchmark, and a unified evaluation protocol, we enable systematic, fair, and reproducible assessment of model performance. 
Using our benchmarks, we establish the first systematic ranking of state-of-the-art layout-guided diffusion models. We also provide separate analyses of text and layout alignment, offering a more interpretable view of model behavior. Fine-grained analyses across scenarios and prompt complexities reveal the strengths and limitations of current approaches. More broadly, our framework supports consistent evaluation and future progress in controllable image generation. 

\section*{Acknowledgments}

We acknowledge ISCRA for awarding this project access to the LEONARDO supercomputer, owned by the EuroHPC Joint Undertaking, hosted by CINECA (Italy) .

{
    \small
    \bibliographystyle{ieeenat_fullname}
    \bibliography{main}
}

\end{document}

%% file: preamble.tex
\usepackage{url}
\usepackage{graphicx}
\usepackage{dutchcal}
\usepackage{wrapfig}
\usepackage{booktabs}
\usepackage{subcaption}
\usepackage{tikz}
\usepackage{makecell}
\usepackage{pifont}  
\usepackage{listings}
\usepackage{longtable}

\newcommand{\cmark}{\ding{51}}%
\newcommand{\xmark}{\ding{55}}%

\newcommand{\csb}{C-Bench}
\newcommand{\osb}{O-Bench}

%% file: main.bib
@String(CVPR= {IEEE Conf. Comput. Vis. Pattern Recog.})

@String(ICCV= {Int. Conf. Comput. Vis.})

@String(ECCV= {Eur. Conf. Comput. Vis.})

@String(ICPR = {Int. Conf. Pattern Recog.})

@String(BMVC= {Brit. Mach. Vis. Conf.})

@String(ICLR = {Int. Conf. Learn. Represent.})

@String(CVPR  = {CVPR})

@String(ICCV  = {ICCV})

@String(ECCV  = {ECCV})

@String(ICPR  = {ICPR})

@String(BMVC  =	{BMVC})

@String(ICLR  = {ICLR})

@inproceedings{DBLP:conf/icpr/ParolariIB24,
  author       = {Luca Parolari and
                  Elena Izzo and
                  Lamberto Ballan},
  title        = {Harlequin: Color-Driven Generation of Synthetic Data for Referring
                  Expression Comprehension},
  booktitle    = {Pattern Recognition - 27th International Conference, {ICPR} 2024,
                  Kolkata, India, December 1-5, 2024, Proceedings, Part {XVIII}},
  series       = {Lecture Notes in Computer Science},
  volume       = {15318},
  pages        = {292--307},
  publisher    = {Springer},
  year         = {2024},
  url          = {https://doi.org/10.1007/978-3-031-78456-9\_19},
  doi          = {10.1007/978-3-031-78456-9\_19},
  bibsource    = {dblp computer science bibliography, https://dblp.org}
}

@inproceedings{DBLP:conf/iccv/BaldratiMCC0C23,
  author       = {Alberto Baldrati and
                  Davide Morelli and
                  Giuseppe Cartella and
                  Marcella Cornia and
                  Marco Bertini and
                  Rita Cucchiara},
  title        = {Multimodal Garment Designer: Human-Centric Latent Diffusion Models
                  for Fashion Image Editing},
  booktitle    = {{IEEE/CVF} International Conference on Computer Vision, {ICCV} 2023,
                  Paris, France, October 1-6, 2023},
  pages        = {23336--23345},
  publisher    = {{IEEE}},
  year         = {2023},
  url          = {https://doi.org/10.1109/ICCV51070.2023.02138},
  doi          = {10.1109/ICCV51070.2023.02138},
  bibsource    = {dblp computer science bibliography, https://dblp.org}
}

@article{izzo20257bench,
  title={7Bench: a Comprehensive Benchmark for Layout-guided Text-to-image Models},
  author={Izzo, Elena and Parolari, Luca and Vezzaro, Davide and Ballan, Lamberto},
  journal={arXiv preprint arXiv:2508.12919},
  year={2025}
}

@inproceedings{DBLP:conf/cvpr/0001LYGWB22,
  author       = {Jaemin Cho and
                  Linjie Li and
                  Zhengyuan Yang and
                  Zhe Gan and
                  Lijuan Wang and
                  Mohit Bansal},
  title        = {Diagnostic Benchmark and Iterative Inpainting for Layout-Guided Image
                  Generation},
  booktitle    = {{IEEE/CVF} Conference on Computer Vision and Pattern Recognition,
                  {CVPR} 2024 - Workshops, Seattle, WA, USA, June 17-18, 2024},
  pages        = {5280--5289},
  publisher    = {{IEEE}},
  year         = {2024},
  url          = {https://doi.org/10.1109/CVPRW63382.2024.00537},
  doi          = {10.1109/CVPRW63382.2024.00537},
  bibsource    = {dblp computer science bibliography, https://dblp.org}
}

@inproceedings{DBLP:conf/nips/HuangSXLL23,
  author       = {Kaiyi Huang and
                  Kaiyue Sun and
                  Enze Xie and
                  Zhenguo Li and
                  Xihui Liu},
  editor       = {Alice Oh and
                  Tristan Naumann and
                  Amir Globerson and
                  Kate Saenko and
                  Moritz Hardt and
                  Sergey Levine},
  title        = {T2I-CompBench: {A} Comprehensive Benchmark for Open-world Compositional
                  Text-to-image Generation},
  booktitle    = {Advances in Neural Information Processing Systems 36: Annual Conference
                  on Neural Information Processing Systems 2023, NeurIPS 2023, New Orleans,
                  LA, USA, December 10 - 16, 2023},
  year         = {2023},
  url          = {http://papers.nips.cc/paper\_files/paper/2023/hash/f8ad010cdd9143dbb0e9308c093aff24-Abstract-Datasets\_and\_Benchmarks.html},
  bibsource    = {dblp computer science bibliography, https://dblp.org}
}

@inproceedings{DBLP:conf/iccv/HuLKWOKS23,
  author       = {Yushi Hu and
                  Benlin Liu and
                  Jungo Kasai and
                  Yizhong Wang and
                  Mari Ostendorf and
                  Ranjay Krishna and
                  Noah A. Smith},
  title        = {{TIFA:} Accurate and Interpretable Text-to-Image Faithfulness Evaluation
                  with Question Answering},
  booktitle    = {{IEEE/CVF} International Conference on Computer Vision, {ICCV} 2023,
                  Paris, France, October 1-6, 2023},
  pages        = {20349--20360},
  publisher    = {{IEEE}},
  year         = {2023},
  url          = {https://doi.org/10.1109/ICCV51070.2023.01866},
  doi          = {10.1109/ICCV51070.2023.01866},
  bibsource    = {dblp computer science bibliography, https://dblp.org}
}

@inproceedings{DBLP:conf/iccv/Bakr0SKLE23,
  author       = {Eslam Mohamed Bakr and
                  Pengzhan Sun and
                  Xiaoqian Shen and
                  Faizan Farooq Khan and
                  Li Erran Li and
                  Mohamed Elhoseiny},
  title        = {HRS-Bench: Holistic, Reliable and Scalable Benchmark for Text-to-Image
                  Models},
  booktitle    = {{IEEE/CVF} International Conference on Computer Vision, {ICCV} 2023,
                  Paris, France, October 1-6, 2023},
  pages        = {19984--19996},
  publisher    = {{IEEE}},
  year         = {2023},
  url          = {https://doi.org/10.1109/ICCV51070.2023.01834},
  doi          = {10.1109/ICCV51070.2023.01834},
  bibsource    = {dblp computer science bibliography, https://dblp.org}
}

@inproceedings{DBLP:conf/nips/SahariaCSLWDGLA22,
  author       = {Chitwan Saharia and
                  William Chan and
                  Saurabh Saxena and
                  Lala Li and
                  Jay Whang and
                  Emily L. Denton and
                  and others},
  title        = {Photorealistic Text-to-Image Diffusion Models with Deep Language Understanding},
  booktitle    = {Advances in Neural Information Processing Systems 35: Annual Conference
                  on Neural Information Processing Systems 2022, NeurIPS 2022, New Orleans,
                  LA, USA, November 28 - December 9, 2022},
  year         = {2022},
  url          = {http://papers.nips.cc/paper\_files/paper/2022/hash/ec795aeadae0b7d230fa35cbaf04c041-Abstract-Conference.html},
  bibsource    = {dblp computer science bibliography, https://dblp.org}
}

@inproceedings{DBLP:conf/cvpr/LiLWMYGLL23,
  author       = {Yuheng Li and
                  Haotian Liu and
                  Qingyang Wu and
                  Fangzhou Mu and
                  Jianwei Yang and
                  Jianfeng Gao and
                  Chunyuan Li and
                  Yong Jae Lee},
  title        = {{GLIGEN:} Open-Set Grounded Text-to-Image Generation},
  booktitle    = {{IEEE/CVF} Conference on Computer Vision and Pattern Recognition,
                  {CVPR} 2023, Vancouver, BC, Canada, June 17-24, 2023},
  pages        = {22511--22521},
  publisher    = {{IEEE}},
  year         = {2023},
  url          = {https://doi.org/10.1109/CVPR52729.2023.02156},
  doi          = {10.1109/CVPR52729.2023.02156},
  bibsource    = {dblp computer science bibliography, https://dblp.org}
}

@inproceedings{DBLP:conf/iccv/XieLHLZ0S23,
  author       = {Jinheng Xie and
                  Yuexiang Li and
                  Yawen Huang and
                  Haozhe Liu and
                  Wentian Zhang and
                  Yefeng Zheng and
                  Mike Zheng Shou},
  title        = {BoxDiff: Text-to-Image Synthesis with Training-Free Box-Constrained
                  Diffusion},
  booktitle    = {{IEEE/CVF} International Conference on Computer Vision, {ICCV} 2023,
                  Paris, France, October 1-6, 2023},
  pages        = {7418--7427},
  publisher    = {{IEEE}},
  year         = {2023},
  url          = {https://doi.org/10.1109/ICCV51070.2023.00685},
  doi          = {10.1109/ICCV51070.2023.00685},
  bibsource    = {dblp computer science bibliography, https://dblp.org}
}

@inproceedings{DBLP:conf/cvpr/PhungGH24,
  author       = {Quynh Phung and
                  Songwei Ge and
                  Jia{-}Bin Huang},
  title        = {Grounded Text-to-Image Synthesis with Attention Refocusing},
  booktitle    = {{IEEE/CVF} Conference on Computer Vision and Pattern Recognition,
                  {CVPR} 2024, Seattle, WA, USA, June 16-22, 2024},
  pages        = {7932--7942},
  publisher    = {{IEEE}},
  year         = {2024},
  url          = {https://doi.org/10.1109/CVPR52733.2024.00758},
  doi          = {10.1109/CVPR52733.2024.00758},
  bibsource    = {dblp computer science bibliography, https://dblp.org}
}

@inproceedings{DBLP:conf/wacv/ChenLV24,
  author       = {Minghao Chen and
                  Iro Laina and
                  Andrea Vedaldi},
  title        = {Training-Free Layout Control with Cross-Attention Guidance},
  booktitle    = {{IEEE/CVF} Winter Conference on Applications of Computer Vision, {WACV}
                  2024, Waikoloa, HI, USA, January 3-8, 2024},
  pages        = {5331--5341},
  publisher    = {{IEEE}},
  year         = {2024},
  url          = {https://doi.org/10.1109/WACV57701.2024.00526},
  doi          = {10.1109/WACV57701.2024.00526},
  bibsource    = {dblp computer science bibliography, https://dblp.org}
}

@inproceedings{DBLP:conf/cvpr/ZhouLMZY24,
  author       = {Dewei Zhou and
                  You Li and
                  Fan Ma and
                  Xiaoting Zhang and
                  Yi Yang},
  title        = {{MIGC:} Multi-Instance Generation Controller for Text-to-Image Synthesis},
  booktitle    = {{IEEE/CVF} Conference on Computer Vision and Pattern Recognition,
                  {CVPR} 2024, Seattle, WA, USA, June 16-22, 2024},
  pages        = {6818--6828},
  publisher    = {{IEEE}},
  year         = {2024},
  url          = {https://doi.org/10.1109/CVPR52733.2024.00651},
  doi          = {10.1109/CVPR52733.2024.00651},
  bibsource    = {dblp computer science bibliography, https://dblp.org}
}

@inproceedings{DBLP:conf/cvpr/RombachBLEO22,
  author       = {Robin Rombach and
                  Andreas Blattmann and
                  Dominik Lorenz and
                  Patrick Esser and
                  Bj{\"{o}}rn Ommer},
  title        = {High-Resolution Image Synthesis with Latent Diffusion Models},
  booktitle    = {{IEEE/CVF} Conference on Computer Vision and Pattern Recognition,
                  {CVPR} 2022, New Orleans, LA, USA, June 18-24, 2022},
  pages        = {10674--10685},
  publisher    = {{IEEE}},
  year         = {2022},
  url          = {https://doi.org/10.1109/CVPR52688.2022.01042},
  doi          = {10.1109/CVPR52688.2022.01042},
  bibsource    = {dblp computer science bibliography, https://dblp.org}
}

@inproceedings{DBLP:conf/bmvc/RigoniPSSB23,
  author       = {Davide Rigoni and
                  Luca Parolari and
                  Luciano Serafini and
                  Alessandro Sperduti and
                  Lamberto Ballan},
  title        = {Weakly-Supervised Visual-Textual Grounding with Semantic Prior Refinement},
  booktitle    = {34th British Machine Vision Conference 2023, {BMVC} 2023, Aberdeen,
                  UK, November 20-24, 2023},
  pages        = {229},
  publisher    = {{BMVA} Press},
  year         = {2023},
  url          = {http://proceedings.bmvc2023.org/229/},
  bibsource    = {dblp computer science bibliography, https://dblp.org}
}

@article{DBLP:journals/ijcv/PlummerWCCHL17,
  author       = {Bryan A. Plummer and
                  Liwei Wang and
                  Chris M. Cervantes and
                  Juan C. Caicedo and
                  Julia Hockenmaier and
                  Svetlana Lazebnik},
  title        = {Flickr30k Entities: Collecting Region-to-Phrase Correspondences for
                  Richer Image-to-Sentence Models},
  journal      = {Int. J. Comput. Vis.},
  volume       = {123},
  number       = {1},
  pages        = {74--93},
  year         = {2017},
  url          = {https://doi.org/10.1007/s11263-016-0965-7},
  doi          = {10.1007/S11263-016-0965-7},
  bibsource    = {dblp computer science bibliography, https://dblp.org}
}

@article{hurst2024gpt,
  title={Gpt-4o system card},
  author={Hurst, Aaron and Lerer, Adam and Goucher, Adam P and Perelman, Adam and Ramesh, Aditya and Clark, Aidan and Ostrow, AJ and Welihinda, Akila and Hayes, Alan and Radford, Alec and others},
  journal={arXiv preprint arXiv:2410.21276},
  year={2024}
}

@inproceedings{
minderer2023scaling,
title={Scaling Open-Vocabulary Object Detection},
author={Matthias Minderer and Alexey A. Gritsenko and Neil Houlsby},
booktitle={Thirty-seventh Conference on Neural Information Processing Systems},
year={2023},
url={https://openreview.net/forum?id=mQPNcBWjGc}
}

@inproceedings{DBLP:conf/iccv/TragerPZABS23,
  author       = {Matthew Trager and
                  Pramuditha Perera and
                  Luca Zancato and
                  Alessandro Achille and
                  Parminder Bhatia and
                  Stefano Soatto},
  title        = {Linear Spaces of Meanings: Compositional Structures in Vision-Language
                  Models},
  booktitle    = {{IEEE/CVF} International Conference on Computer Vision, {ICCV} 2023, Paris, France, October 1-6, 2023},
  year         = {2023},
}

@misc{spacy2020,
  author       = {Honnibal, Matthew and Montani, Ines and Van Landeghem, Sofie and Boyd, Adriane},
  title        = {spaCy: Industrial-strength Natural Language Processing in Python},
  year         = {2020},
  doi          = {10.5281/zenodo.1212303},
  url          = {https://doi.org/10.5281/zenodo.1212303},
}

@article{DBLP:journals/jasis/Blair79,
  author       = {David C. Blair},
  title        = {Information Retrieval, 2nd ed. {C.J.} Van Rijsbergen. London: Butterworths;
                  1979: 208 pp. Price: {\textdollar}32.50},
  journal      = {J. Am. Soc. Inf. Sci.},
  volume       = {30},
  number       = {6},
  pages        = {374--375},
  year         = {1979},
  url          = {https://doi.org/10.1002/asi.4630300621},
  doi          = {10.1002/ASI.4630300621},
  bibsource    = {dblp computer science bibliography, https://dblp.org}
}

@inproceedings{DBLP:conf/wacv/GrimalBFT24,
  author       = {Paul Grimal and
                  Herv{\'{e}} Le Borgne and
                  Olivier Ferret and
                  Julien Tourille},
  title        = {{TIAM} - {A} Metric for Evaluating Alignment in Text-to-Image Generation},
  booktitle    = {{IEEE/CVF} Winter Conference on Applications of Computer Vision, {WACV}
                  2024, Waikoloa, HI, USA, 2024},
  pages        = {2878--2887},
  publisher    = {{IEEE}},
  year         = {2024},
  url          = {https://doi.org/10.1109/WACV57701.2024.00287},
  doi          = {10.1109/WACV57701.2024.00287},
  bibsource    = {dblp computer science bibliography, https://dblp.org}
}

@inproceedings{DBLP:conf/nips/BrownMRSKDNSSAA20,
  author       = {Tom B. Brown and
                  Benjamin Mann and
                  Nick Ryder and
                  Melanie Subbiah and
                  Jared Kaplan and
                  and others},
  editor       = {Hugo Larochelle and
                  Marc'Aurelio Ranzato and
                  Raia Hadsell and
                  Maria{-}Florina Balcan and
                  Hsuan{-}Tien Lin},
  title        = {Language Models are Few-Shot Learners},
  booktitle    = {Advances in Neural Information Processing Systems 33: NeurIPS 2020, 2020, virtual},
  year         = {2020},
  url          = {https://proceedings.neurips.cc/paper/2020/hash/1457c0d6bfcb4967418bfb8ac142f64a-Abstract.html},
  bibsource    = {dblp computer science bibliography, https://dblp.org}
}

@article{DBLP:journals/corr/abs-2307-09288,
  author={Touvron, Hugo and Martin, Louis and Stone, Kevin and Albert, Peter and Almahairi, Amjad and Babaei, Yasmine and Bashlykov, Nikolay and Batra, Soumya and Bhargava, Prajjwal and Bhosale, Shruti and others},
  title        = {Llama 2: Open Foundation and Fine-Tuned Chat Models},
  journal      = {CoRR},
  volume       = {abs/2307.09288},
  year         = {2023},
  url          = {https://doi.org/10.48550/arXiv.2307.09288},
  doi          = {10.48550/ARXIV.2307.09288},
  eprinttype    = {arXiv},
  eprint       = {2307.09288},
  bibsource    = {dblp computer science bibliography, https://dblp.org}
}

@inproceedings{DBLP:conf/emnlp/KhashabiMKSTCH20,
  author       = {Daniel Khashabi and
                  Sewon Min and
                  Tushar Khot and
                  Ashish Sabharwal and
                  Oyvind Tafjord and
                  Peter Clark and
                  Hannaneh Hajishirzi},
  editor       = {Trevor Cohn and
                  Yulan He and
                  Yang Liu},
  title        = {UnifiedQA: Crossing Format Boundaries With a Single {QA} System},
  booktitle    = {Findings of the Association for Computational Linguistics: {EMNLP}
                  2020, Online Event, 16-20 November 2020},
  series       = {Findings of {ACL}},
  volume       = {{EMNLP} 2020},
  pages        = {1896--1907},
  publisher    = {Association for Computational Linguistics},
  year         = {2020},
  url          = {https://doi.org/10.18653/v1/2020.findings-emnlp.171},
  doi          = {10.18653/V1/2020.FINDINGS-EMNLP.171},
  bibsource    = {dblp computer science bibliography, https://dblp.org}
}

@inproceedings{DBLP:conf/icml/0008LSH23,
  author       = {Junnan Li and
                  Dongxu Li and
                  Silvio Savarese and
                  Steven C. H. Hoi},
  editor       = {Andreas Krause and
                  Emma Brunskill and
                  Kyunghyun Cho and
                  Barbara Engelhardt and
                  Sivan Sabato and
                  Jonathan Scarlett},
  title        = {{BLIP-2:} Bootstrapping Language-Image Pre-training with Frozen Image
                  Encoders and Large Language Models},
  booktitle    = {International Conference on Machine Learning, {ICML} 2023, 23-29 July
                  2023, Honolulu, Hawaii, {USA}},
  series       = {Proceedings of Machine Learning Research},
  volume       = {202},
  pages        = {19730--19742},
  publisher    = {{PMLR}},
  year         = {2023},
  url          = {https://proceedings.mlr.press/v202/li23q.html},
  bibsource    = {dblp computer science bibliography, https://dblp.org}
}

@inproceedings{DBLP:conf/iccv/DengYCZL21,
  author       = {Jiajun Deng and
                  Zhengyuan Yang and
                  Tianlang Chen and
                  Wengang Zhou and
                  Houqiang Li},
  title        = {TransVG: End-to-End Visual Grounding with Transformers},
  booktitle    = {2021 {IEEE/CVF} International Conference on Computer Vision, {ICCV}
                  2021, Montreal, QC, Canada, October 10-17, 2021},
  pages        = {1749--1759},
  publisher    = {{IEEE}},
  year         = {2021},
  url          = {https://doi.org/10.1109/ICCV48922.2021.00179},
  doi          = {10.1109/ICCV48922.2021.00179},
  bibsource    = {dblp computer science bibliography, https://dblp.org}
}

@article{DBLP:journals/corr/abs-2211-01324,
  author       = {Yogesh Balaji and
                  Seungjun Nah and
                  Xun Huang and
                  Arash Vahdat and
                  Jiaming Song and
                  Karsten Kreis and
                  Miika Aittala and
                  Timo Aila and
                  Samuli Laine and
                  Bryan Catanzaro and
                  Tero Karras and
                  Ming{-}Yu Liu},
  title        = {eDiff-I: Text-to-Image Diffusion Models with an Ensemble of Expert
                  Denoisers},
  journal      = {CoRR},
  volume       = {abs/2211.01324},
  year         = {2022},
  url          = {https://doi.org/10.48550/arXiv.2211.01324},
  doi          = {10.48550/ARXIV.2211.01324},
  eprinttype    = {arXiv},
  eprint       = {2211.01324},
  bibsource    = {dblp computer science bibliography, https://dblp.org}
}

@inproceedings{DBLP:conf/nips/HoJA20,
  author       = {Jonathan Ho and
                  Ajay Jain and
                  Pieter Abbeel},
  editor       = {Hugo Larochelle and
                  Marc'Aurelio Ranzato and
                  Raia Hadsell and
                  Maria{-}Florina Balcan and
                  Hsuan{-}Tien Lin},
  title        = {Denoising Diffusion Probabilistic Models},
  booktitle    = {Advances in Neural Information Processing Systems 33: Annual Conference
                  on Neural Information Processing Systems 2020, NeurIPS 2020, December
                  6-12, 2020, virtual},
  year         = {2020},
  url          = {https://proceedings.neurips.cc/paper/2020/hash/4c5bcfec8584af0d967f1ab10179ca4b-Abstract.html},
  bibsource    = {dblp computer science bibliography, https://dblp.org}
}

@article{DBLP:journals/corr/abs-2112-10741,
  author       = {Alex Nichol and
                  Prafulla Dhariwal and
                  Aditya Ramesh and
                  Pranav Shyam and
                  Pamela Mishkin and
                  Bob McGrew and
                  Ilya Sutskever and
                  Mark Chen},
  title        = {{GLIDE:} Towards Photorealistic Image Generation and Editing with
                  Text-Guided Diffusion Models},
  journal      = {CoRR},
  volume       = {abs/2112.10741},
  year         = {2021},
  url          = {https://arxiv.org/abs/2112.10741},
  eprinttype    = {arXiv},
  eprint       = {2112.10741},
  bibsource    = {dblp computer science bibliography, https://dblp.org}
}

@inproceedings{DBLP:conf/iclr/FengHFJANBWW23,
  author       = {Weixi Feng and
                  Xuehai He and
                  Tsu{-}Jui Fu and
                  Varun Jampani and
                  Arjun R. Akula and
                  Pradyumna Narayana and
                  Sugato Basu and
                  Xin Eric Wang and
                  William Yang Wang},
  title        = {Training-Free Structured Diffusion Guidance for Compositional Text-to-Image
                  Synthesis},
  booktitle    = {The Eleventh International Conference on Learning Representations,
                  {ICLR} 2023, Kigali, Rwanda, May 1-5, 2023},
  publisher    = {OpenReview.net},
  year         = {2023},
  url          = {https://openreview.net/forum?id=PUIqjT4rzq7},
  bibsource    = {dblp computer science bibliography, https://dblp.org}
}

@inproceedings{DBLP:conf/nips/WuYHRA24,
  title = {{{ConceptMix}}: {{A}} Compositional Image Generation Benchmark with Controllable Difficulty},
  booktitle = {Advances in Neural Information Processing Systems 38: {{Annual}} Conference on Neural Information Processing Systems 2024, {{NeurIPS}} 2024, Vancouver, {{BC}}, Canada, December 10 - 15, 2024},
  author = {Wu, Xindi and Yu, Dingli and Huang, Yangsibo and Russakovsky, Olga and Arora, Sanjeev},
  editor = {Globersons, Amir and Mackey, Lester and Belgrave, Danielle and Fan, Angela and Paquet, Ulrich and Tomczak, Jakub M. and Zhang, Cheng},
  date = {2024},
  year = {2024},
  url = {http://papers.nips.cc/paper_files/paper/2024/hash/9c3563bbeb2ad7f3b3b8ed0fcd3b440f-Abstract-Datasets_and_Benchmarks_Track.html},
  bibsource = {dblp computer science bibliography, https://dblp.org}
}

@article{DBLP:journals/corr/abs-2412-03859,
  title = {{{CreatiLayout}}: {{Siamese}} Multimodal Diffusion Transformer for Creative Layout-to-Image Generation},
  author = {Zhang, Hui and Hong, Dexiang and Gao, Tingwei and Wang, Yitong and Shao, Jie and Wu, Xinglong and Wu, Zuxuan and Jiang, Yu-Gang},
  date = {2024},
  year = {2024},
  journal = {CoRR},
  volume = {abs/2412.03859},
  eprint = {2412.03859},
  eprinttype = {arXiv},
  doi = {10.48550/ARXIV.2412.03859},
  url = {https://doi.org/10.48550/arXiv.2412.03859},
  bibsource = {dblp computer science bibliography, https://dblp.org}
}

@inproceedings{DBLP:conf/nips/ChengMwLMWLY24,
  title = {{{HiCo}}: {{Hierarchical}} Controllable Diffusion Model for Layout-to-Image Generation},
  booktitle = {Advances in Neural Information Processing Systems 38: {{Annual}} Conference on Neural Information Processing Systems 2024, {{NeurIPS}} 2024, Vancouver, {{BC}}, Canada, December 10 - 15, 2024},
  author = {Cheng, Bo and Ma, Yuhang and Wu, Liebucha and Liu, Shanyuan and Ma, Ao and Wu, Xiaoyu and Leng, Dawei and Yin, Yuhui},
  editor = {Globersons, Amir and Mackey, Lester and Belgrave, Danielle and Fan, Angela and Paquet, Ulrich and Tomczak, Jakub M. and Zhang, Cheng},
  date = {2024},
  year = {2024},
  url = {http://papers.nips.cc/paper_files/paper/2024/hash/e8da5e7ed327823326360ac3c7d7f833-Abstract-Conference.html},
  bibsource = {dblp computer science bibliography, https://dblp.org}
}

@inproceedings{DBLP:conf/iclr/GaniBN0W24,
  title = {{{LLM}} Blueprint: {{Enabling}} Text-to-Image Generation with Complex and Detailed Prompts},
  booktitle = {The Twelfth International Conference on Learning Representations, {{ICLR}} 2024, Vienna, Austria, May 7-11, 2024},
  author = {Gani, Hanan and Bhat, Shariq Farooq and Naseer, Muzammal and Khan, Salman and Wonka, Peter},
  date = {2024},
  year = {2024},
  publisher = {OpenReview.net},
  url = {https://openreview.net/forum?id=mNYF0IHbRy},
  bibsource = {dblp computer science bibliography, https://dblp.org}
}

@article{DBLP:journals/tmlr/LianLYD24,
  title = {{{LLM-grounded}} Diffusion: {{Enhancing}} Prompt Understanding of Text-to-Image Diffusion Models with Large Language Models},
  author = {Lian, Long and Li, Boyi and Yala, Adam and Darrell, Trevor},
  date = {2024},
  year = {2024},
  journal = {Trans. Mach. Learn. Res.},
  volume = {2024},
  url = {https://openreview.net/forum?id=hFALpTb4fR},
  bibsource = {dblp computer science bibliography, https://dblp.org}
}

@inproceedings{DBLP:conf/iccv/LiWKTS21,
  author       = {Zejian Li and
                  Jingyu Wu and
                  Immanuel Koh and
                  Yongchuan Tang and
                  Lingyun Sun},
  title        = {Image Synthesis from Layout with Locality-Aware Mask Adaption},
  booktitle    = ICCV,
  year         = {2021},
}

@inproceedings{DBLP:conf/sigir/Robertson08,
  author       = {Stephen Robertson},
  editor       = {Sung{-}Hyon Myaeng and
                  Douglas W. Oard and
                  Fabrizio Sebastiani and
                  Tat{-}Seng Chua and
                  Mun{-}Kew Leong},
  title        = {A new interpretation of average precision},
  booktitle    = {Proceedings of the 31st Annual International {ACM} {SIGIR} Conference
                  on Research and Development in Information Retrieval, {SIGIR} 2008,
                  Singapore, July 20-24, 2008},
  pages        = {689--690},
  publisher    = {{ACM}},
  year         = {2008},
  url          = {https://doi.org/10.1145/1390334.1390453},
  doi          = {10.1145/1390334.1390453},
  bibsource    = {dblp computer science bibliography, https://dblp.org}
}

@inproceedings{DBLP:conf/eccv/LinMBHPRDZ14,
  author       = {Tsung{-}Yi Lin and
                  Michael Maire and
                  Serge J. Belongie and
                  James Hays and
                  Pietro Perona and
                  Deva Ramanan and
                  Piotr Doll{\'{a}}r and
                  C. Lawrence Zitnick},
  editor       = {David J. Fleet and
                  Tom{\'{a}}s Pajdla and
                  Bernt Schiele and
                  Tinne Tuytelaars},
  title        = {Microsoft {COCO:} Common Objects in Context},
  booktitle    = {Computer Vision - {ECCV} 2014 - 13th European Conference, Zurich,
                  Switzerland, September 6-12, 2014, Proceedings, Part {V}},
  series       = {Lecture Notes in Computer Science},
  volume       = {8693},
  pages        = {740--755},
  publisher    = {Springer},
  year         = {2014},
  url          = {https://doi.org/10.1007/978-3-319-10602-1\_48},
  doi          = {10.1007/978-3-319-10602-1\_48},
  bibsource    = {dblp computer science bibliography, https://dblp.org}
}

@article{DBLP:journals/corr/abs-2306-14824,
  author       = {Zhiliang Peng and
                  Wenhui Wang and
                  Li Dong and
                  Yaru Hao and
                  Shaohan Huang and
                  Shuming Ma and
                  Furu Wei},
  title        = {Kosmos-2: Grounding Multimodal Large Language Models to the World},
  journal      = {CoRR},
  volume       = {abs/2306.14824},
  year         = {2023},
  url          = {https://doi.org/10.48550/arXiv.2306.14824},
  doi          = {10.48550/ARXIV.2306.14824},
  eprinttype    = {arXiv},
  eprint       = {2306.14824},
  bibsource    = {dblp computer science bibliography, https://dblp.org}
}

@article{DBLP:journals/corr/abs-2205-06230,
  author       = {Matthias Minderer and
                  Alexey A. Gritsenko and
                  Austin Stone and
                  Maxim Neumann and
                  Dirk Weissenborn and
                  Alexey Dosovitskiy and
                  Aravindh Mahendran and
                  Anurag Arnab and
                  Mostafa Dehghani and
                  Zhuoran Shen and
                  Xiao Wang and
                  Xiaohua Zhai and
                  Thomas Kipf and
                  Neil Houlsby},
  title        = {Simple Open-Vocabulary Object Detection with Vision Transformers},
  journal      = {CoRR},
  volume       = {abs/2205.06230},
  year         = {2022},
  url          = {https://doi.org/10.48550/arXiv.2205.06230},
  doi          = {10.48550/ARXIV.2205.06230},
  eprinttype    = {arXiv},
  eprint       = {2205.06230},
  bibsource    = {dblp computer science bibliography, https://dblp.org}
}
